\documentclass[conference]{IEEEtran}
\IEEEoverridecommandlockouts

\usepackage{cite}
\usepackage{amsmath,amssymb,amsfonts}
\usepackage{algorithmic}
\usepackage{graphicx}
\usepackage{textcomp}
\usepackage{xcolor}

\usepackage{todonotes}
\usepackage{bm} 
\usepackage{amsthm} 
\usepackage{microtype}

\newtheorem{definition}{Definition}

\iftrue
    \newcommand{\aneta}[1]{\textcolor{green}{Aneta: #1}}
    \newcommand{\jendrik}[1]{\textcolor{orange}{Jendrik: #1}}
    \newcommand{\kristina}[1]{\textcolor{cyan}{Kristina: #1}}
    \newcommand{\nikos}[1]{\textcolor{red}{Nikos: #1}}
    \newcommand{\thanos}[1]{\textcolor{yellow}{Thanos: #1}}
\else
    \newcommand{\aneta}[1]{}
    \newcommand{\jendrik}[1]{}
    \newcommand{\kristina}[1]{}
    \newcommand{\nikos}[1]{}
    \newcommand{\thanos}[1]{}
\fi

\newcommand{\set}[1]{\{#1\}}

\newcommand{\props}{\ensuremath{{\mathcal{P}}}}
\newcommand{\psa}{\ensuremath{{\props\mathit{SA}}}}

\newcommand{\MP}{\ensuremath{\mathcal{P}}}
\newcommand{\MR}{\ensuremath{\mathcal{M}}}

\def\BibTeX{{\rm B\kern-.05em{\sc i\kern-.025em b}\kern-.08em
    T\kern-.1667em\lower.7ex\hbox{E}\kern-.125emX}}
\begin{document}

\title{Reinforcement Learning with Reward Machines for Sleep Control in Mobile Networks \\
\thanks{This work was supported by Ericsson Research and the Wallenberg AI, Autonomous Systems, and Software Program (WASP) funded by the Knut and Alice Wallenberg Foundation. The work of N. Pappas has been supported in part by ELLIIT and the European Union (6G-LEADER, 101192080).}
}

\author{
	\IEEEauthorblockN{Kristina Levina\IEEEauthorrefmark{1}\IEEEauthorrefmark{2}, Nikolaos Pappas\IEEEauthorrefmark{1}, Athanasios Karapantelakis\IEEEauthorrefmark{2}, Aneta Vulgarakis Feljan\IEEEauthorrefmark{2}, and Jendrik Seipp\IEEEauthorrefmark{1}}
	\IEEEauthorblockA{\IEEEauthorrefmark{1}Department of Computer and Information Science, Link\"{o}ping University, Link\"{o}ping, Sweden}
	\IEEEauthorblockA{\IEEEauthorrefmark{2}Ericsson Research}
	\{kristina.levina, nikolaos.pappas, jendrik.seipp\}@liu.se, \{athanasios.karapantelakis, aneta.vulgarakis\}@ericsson.com
	}

\maketitle

\begin{abstract}
Energy efficiency in mobile networks is crucial for sustainable telecommunications infrastructure, particularly as network densification continues to increase power consumption. 
Sleep mechanisms for the components in mobile networks can reduce energy use, but deciding which components to put to sleep, when, and for how long while preserving quality of service (QoS) remains a difficult optimisation problem. 
In this paper, we utilise reinforcement learning with reward machines (RMs) to make sleep-control decisions that balance immediate energy savings and long-term QoS impact---time-averaged packet drop rates for deadline-constrained traffic and time-averaged minimum-throughput guarantees for constant-rate users. 
A challenge is that time-averaged constraints depend on cumulative performance over time rather than immediate performance. 
As a result, the effective reward is non-Markovian, and optimal actions depend on operational history rather than the instantaneous system state. 
RMs account for the history dependence by maintaining an abstract state that explicitly tracks the QoS constraint violations over time. 
Our framework provides a principled, scalable approach to energy management for next-generation mobile networks under diverse traffic patterns and QoS requirements.
\end{abstract}


\section{Introduction}

Energy consumption in telecommunications infrastructure has become a critical concern as networks expand to meet increasing data demands through densification~\cite{lopez2022survey}. 
Radio base stations (RBSs) account for the majority of network energy consumption, largely due to the significant power consumption of their components even under low traffic conditions~\cite{auer2011much}. 
Sleep mode (SM) mechanisms reduce energy consumption by dynamically transitioning RBS components into low-power states during periods of low traffic demand~\cite{imran2012infso}.

In this paper, we study the problem of optimising the sleep control of radio units (RUs). 
Modern RUs support multiple SMs, each with distinct power consumption, sleep duration, and wake-up energy cost~\cite{3gpp_rel18_energy}. 
Deciding which RUs to put to sleep, when, and for how long, while maintaining quality-of-service (QoS) guarantees, is a challenging control problem. 
In particular, we must balance immediate energy savings against time-averaged QoS constraints, including packet drop rates for deadline-constrained traffic and minimum throughput for constant-rate users. 
Uncertainties in the wireless environment amplify the challenge: temporally correlated channel conditions, stochastic traffic arrivals, and dynamic user demands.

State-of-the-art stochastic optimisation techniques~\cite{neely2010stochastic, TIT2012}, e.g., Lyapunov optimisation, have been widely used to handle time-averaged constraints in wireless networks. 
By transforming long-term constraints into virtual-queue-stability problems, these methods guarantee asymptotic optimality and enable online control without requiring prior knowledge of traffic or channel statistics. 
However, Lyapunov-based methods can face scalability challenges, as they require solving a per-slot optimisation problem that may be computationally complex (e.g., mixed-integer or non-convex), particularly when the action space is large~\cite{moltafet2021power,taleb2022reliable}. 
This limitation becomes pronounced in the SM selection problem, where multiple RUs must be jointly controlled, leading to an exponential growth in the action space.

Another state-of-the-art approach is constrained Markov decision processes (CMDPs), where optimal policies can be characterised as randomised stationary policies~\cite{altman2021constrained}.
Such policies define a fixed distribution over actions conditioned on the current state and achieve optimality asymptotically.
However, they are inherently memoryless and do not account for temporal correlations. 

Energy-efficient operation via sleep mechanisms has also been studied using analytical models. For instance, in~\cite{jiang2017optimal}, optimal sleeping policies are derived for multiple servers under Markov-modulated Poisson process traffic using an MDP framework.
While such approaches yield structured optimal policies under the assumed stochastic model, they require full knowledge of system dynamics and traffic statistics. In addition, their reliance on explicit modeling limits their adaptability and scalability in complex or high-dimensional settings.

Reinforcement learning (RL) offers a scalable alternative for high-dimensional problems. Recent works have explored hybrid approaches that combine Lyapunov optimisation with deep RL \cite{LuoTCOM25}.
More broadly, constrained RL methods, including constrained policy optimisation~\cite{achiam2017constrained} and Lagrangian (primal--dual) approaches~\cite{stooke2020responsive}, explicitly incorporate constraints into policy learning.
However, these methods typically assume Markovian reward structures and may struggle to capture temporally extended objectives and multi-slot commitments.

To address these limitations, we propose combining RL with reward machines (RMs)~\cite{icarte2022reward}.
RMs provide a structured representation of non-Markovian rewards via a finite-state automaton that tracks progress toward temporally extended objectives. 
In particular, we represent each QoS constraint through an RM---explicit finite-state memory that records the history of constraint violations. 
By augmenting the system state with the RM state, the problem becomes Markovian while preserving the temporal structure. 
This enables efficient learning of policies that handle multi-slot commitments and long-term QoS constraints, making the approach well-suited for SM selection in dynamic wireless environments.

\section{System Model}

We consider a cellular network consisting of a single RBS equipped with $G$ individual RUs. 
Each RU $g \in \mathcal{G} = \set{1, 2, \dots, G}$ can operate in one active and $H$ sleep modes indexed by $h \in \mathcal{H} = \set{0, 1, 2, ..., H}$, each with a duration $\Delta^h$ and a switching latency $\Delta_{\text{sw}}^h$. 
Time is slotted, and each time slot $t \in \mathbb{N}$ has a fixed duration of $\Delta$. 
When active, an idle RU still consumes power $W^{0}$. 
In any SM $h \neq 0$, the power consumption reduces to $W^{h} < W^{0}$. 
The transition from SM $h$ to the active mode incurs a switching energy cost $E_{\text{sw}}^h$.

\subsubsection{Network Topology}

\begin{figure}[h]
\centering
\includegraphics[width=0.4\textwidth]{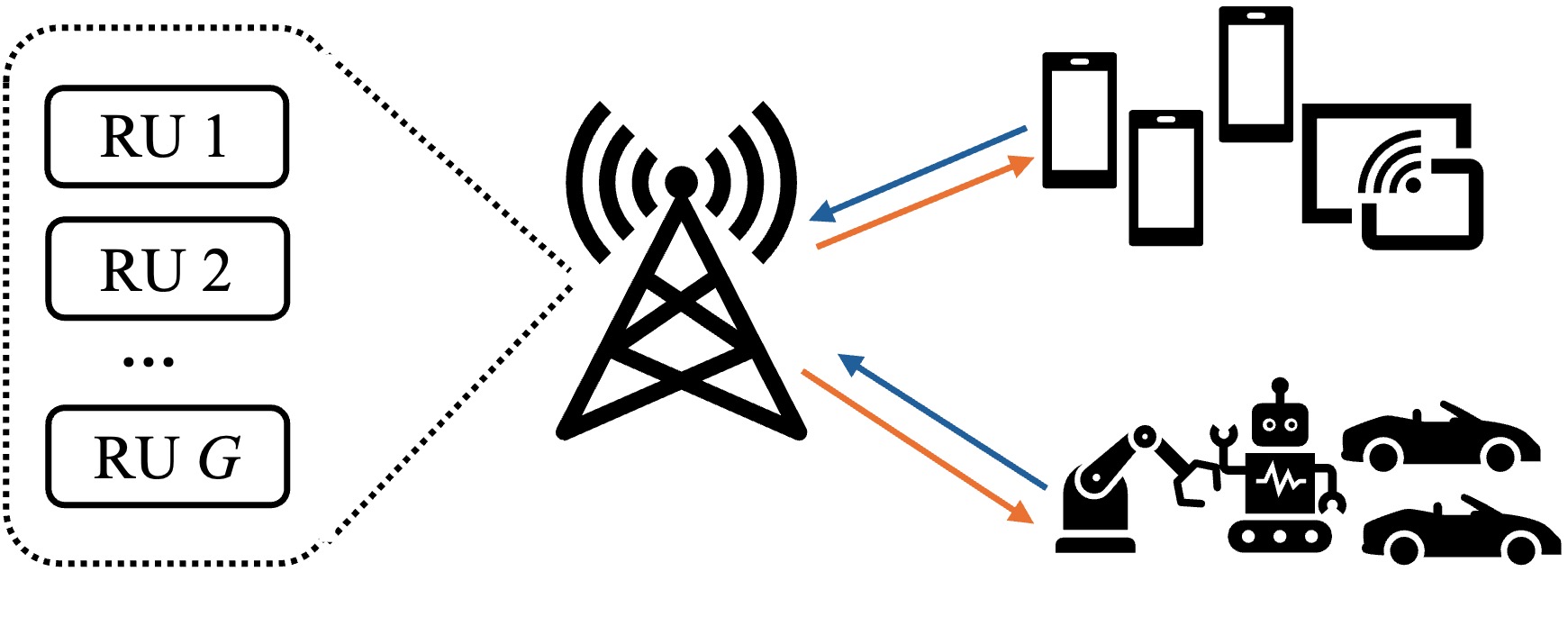}
\caption{One RBS with $G$ radio units (RUs) serving heterogeneous traffic.}
\label{fig:system}
\end{figure}

In the system, $N$ users communicate with the single RBS over wireless fading links (one link per user).
Let $\mathcal{N} = \set{1, \dots, N}$ be the set of all users. 
At each time slot $t$, a central controller dynamically decides which RUs to put into sleep and for how long. 
Sleeping RUs wake themselves up after the sleep duration has elapsed. Formally, the decision is made about the state evolution $h^g_t \in \mathcal{H}$ of each RU $g$. 

We consider two sets of users: users with constant-rate traffic $\mathcal{N}^{m} \subseteq \mathcal{N}$ and users transmitting deadline-constrained packets $\mathcal{N}^{d} \subseteq \mathcal{N}$, such that $\mathcal{N}^{m} \cup \mathcal{N}^{d} = \mathcal{N}$ and $\mathcal{N}^{m} \cap \mathcal{N}^{d} = \emptyset$.
Users in $\mathcal{N}^{m}$ require a minimum average throughput.
For users in $\mathcal{N}^{d}$, a packet is dropped and removed from the system upon deadline expiration.
For user $i^d \in \mathcal{N}^{d}$, the packet deadlines are equal and are denoted by $T^{i^d}\in \mathbb{N}$.
Each user has an associated queue $i\in \mathcal{N}$ with a finite buffer size $B$.
In each queue, packets are served in first-in-first-out (FIFO) order, and no collisions are allowed.
Any RU can serve any queue, but users in $\mathcal{N}^{d}$ are served first.
The packet arrival process is $\bm{\alpha}_t = [\alpha^1_t, \dots, \alpha^N_t]$, where $\alpha^i_t \in \set{0, 1}$ denotes a Bernoulli arrival process.

\subsubsection{Channel Model}

At the beginning of each time slot, the current discrete channel state is observed for each user and is assumed to be accurate while future channel states are unknown.
We assume that the channel state does not change within a time slot but can change between slots.
Let $\bm{Y}_t = [Y^1_t, \dots, Y^N_t]$ represent the channel state vector for each user $i \in \mathcal{N}$ during slot $t$.
We assume two possible channel states $Y^i_t \in \set{0, 1}$: ``Bad'' (deep fading) and ``Good'' (mild fading).
The random variables of the channel process $\bm{Y}_t$ are distributed according to the Gilbert--Elliot model from one slot to the next~\cite{gilbert1960capacity}.

\subsubsection{Traffic Model}

Let $\bm{w}_t = [w^1_t, \dots, w^N_t]$ denote the power allocation vector at $t$.
The set of available power levels is $\set{0, W^{\text{(Low)}}, W^{\text{(High)}}}$, where
$W^{\text{(Low)}}$ and $W^{\text{(High)}}$ are the required powers to have a successful transmission under ``Bad'' and ``Good'' channel conditions, respectively.
Thus,
\begin{equation*}
w^i_t \in \begin{cases}
\{0, W^{\text{(High)}}\}, & \text{if } Y^i_t = 0 \\
\{0, W^{\text{(Low)}}\}, & \text{if } Y^i_t = 1
\end{cases}
, \quad \forall i \in \mathcal{N}.
\end{equation*}

Let $\mu^i_t$ be the data served for user $i$ at $t$.
For each user $i^{d} \in \mathcal{N}^d$, a packet is dropped if its deadline has expired.
Considering the FIFO queue, finite buffer size $B$, and same deadline for all packets in queue $i^{d}$, packets are dropped under the following two conditions:
A packet at the head of the queue is dropped if a new packet arrives at $t$ when the queue length is already $B$; and
all packets in queue $i^{d}$ are dropped if the remaining number of slots to serve the packet is $1$, that is, the deadline $T^{i^{d}}$ expires when $T^{i^{d}}-t=1$.
We denote the dropped data for user $i^{d}$ during time slot $t$ by $\eta^{i^{d}}_t$ and packet drop rate by $D^{i^d}_t$.
Let $O^i_t$ be the number of packets in queue $i$ at $t$.
The queue evolution for each user $i^{d} \in \mathcal{N}^d$ is then 
\begin{equation*}
 O^{i^{d}}_{t + 1} = \max \set{O^{i^{d}}_{t}-\mu^{i^{d}}_{t}, 0} + \alpha^{i^{d}}_{t} - \eta^{i^{d}}_t.
\end{equation*}

We define the average packet drop rate for users $\mathcal{N}^d$ as $\overline{D^{\mathcal{N}^d}} = \lim_{\tau \to \infty} \frac{1}{\tau} \sum_{t=0}^{\tau-1} \sum_{i^{d} \in \mathcal{N}^d}D^{i^{d}}_t$ and average throughput for users $\mathcal{N}^{m}$ as $\overline{\mu^{\mathcal{N}^{m}}} = \lim_{\tau \to \infty} \frac{1}{\tau} \sum_{t=0}^{\tau-1} \sum_{i^{m} \in \mathcal{N}^{m}} \mu^{i^{m}}_{t}$.

\section{Background on Reinforcement Learning with Reward Machines}

In the RL framework, an agent interacts with an environment and receives feedback in the form of rewards. 
The goal is to learn a policy that maximises the total expected reward over time. 
The reward function is typically Markovian. 
RMs are automata that encode temporal information or task-specific objectives. 
Unlike standard reward functions, RMs can handle non-Markovian reward signals. For complex tasks that are difficult to specify in a traditional Markov decision process (MDP), RMs provide the RL agent with memory, improving sample efficiency. 
In telecommunications systems, RMs can help optimise network performance by aligning agent actions with long-term communication objectives and user requirements~\cite{arana2025explainable}.
For a detailed introduction to RL, see~\cite{sutton2018reinforcement}, and for a more complete overview of RMs, see~\cite{icarte2022reward}.

\subsubsection{Reinforcement Learning}

Single-agent RL tasks are generally formalised via MDPs, defined by a tuple $\mathcal{M} = \langle S, s_0, A, p, r, \gamma \rangle$, where $S$ is a finite set of environment states, $s_0 \in S$ is an initial state, $A$ is a finite set of actions, $p(s'|s, a)$ defines the transition probabilities, $r : S \times A \times S \to \mathbb{R}$ is a reward function, and $\gamma \in(0, 1)$ is a discount factor.
A policy $\pi(a|s)$ maps the state space $S$ to the action space $A$.

In state $s_t$, the agent performs action $a_t$ according to policy $\pi(a_t|s_t)$, transitions to state $s_{t+1}$ according to the transition probability $p(s_{t+1}|s_t,a_t)$, and receives reward $r_{t+1}$.
The process repeats until episode termination or reaching a goal state.
The objective is to find an optimal policy $\pi^*(a_t|s_t = s)$ for all $s \in S$ that maximises the expected return $\mathbb{E}_{\pi^*}[ \, \sum^{K-1}_{k = 0}\gamma^k r_{t+k+1}|s_t = s]$, where $K$ is the episode length.
The $Q$-function $q^\pi(s,a)$ quantifies the expected return obtained by taking action $a$ in state $s$ and following policy $\pi$ thereafter. Formally,
$q^\pi(s,a) = \mathbb{E}_{\pi}[ \, \sum^{K-1}_{k = 0}\gamma^k r_{t+k+1}|s_t = s, a_t = a ].$ \,
For an optimal policy $\pi^*$, $q^* = q^{\pi^*}$.

To estimate $q^*(s, a)$ for problems with continuous or high-dimensional state/action spaces, deep RL methods with function approximation are commonly used~\cite{sutton2018reinforcement}.
Twin delayed deep deterministic policy gradient (TD3)~\cite{fujimoto2018addressing} is one such method that combines $Q$-learning with an actor--critic architecture for continuous action spaces.
TD3 employs an actor network for deterministic actions and two critic networks for the $Q$-value estimation.
The ability to handle continuous actions makes TD3 particularly suitable for problems where discrete actions would lead to combinatorial explosion, such as coordinated SM selection across multiple RUs.

\subsubsection{Reward Machines} \label{sect:rms}

An RM is a finite-state machine that represents the reward structure of the environment.
An RM outputs the reward the agent receives upon transitioning between two abstract RM states.

\begin{definition}[Reward machine]
 \label{def:rm-bool}
 An RM is a tuple $\MR^{\text{RM}}_\psa = \langle U, u_0, F, \delta_u, \delta_r \rangle$ given sets of propositional symbols $\MP$, environment states $S$, and actions $A$.
 In the tuple, $U$ is a finite set of states, $u_0$ is an initial state, $F$ is a finite set of terminal states, $\delta_u:U \times 2^\props \to U \cup F$ is a state-transition function, and $\delta_r:U \to [S \times A \times S \to \mathbb{R}]$ is a state-reward function.
\end{definition}

At each time step, the RM receives
the set of propositions that are true in the current environment state.
The transition function then selects the next abstract successor state, and the reward function assigns the corresponding reward.

Intuitively, an MDP with RMs (MDPRM) is an MDP defined over the cross-product $\tilde{S} = S \times (U\cup F)$: a tuple $\MR^{\text{RM}} = \langle \tilde{S}, \tilde{s}_0, \tilde{A}, \tilde{p}, \tilde{r}, \tilde{\gamma} \rangle$, where
$\tilde{s}_0 \in \tilde{S}$ is an initial state;
$\tilde{A} = A$ is a set of actions; 
state-transition function $\tilde{p}(\langle s', u'\rangle | \langle s, u\rangle, a)$ is $p(s'| s, a)$ if $u' = \delta_u(u, \nu(s, a, s'))$ (where $\nu$ is a labelling function) and $u \in U$, $p(s'| s, a)$ if $u' = u$ and $u \in F$, and $0$ otherwise;
state-reward function $\tilde{r}(\langle s, u\rangle, a, \langle s', u'\rangle)$ is $\delta_r(u)(s, a, s')$ if $u \notin  F$ and $0$ otherwise;
and $\tilde{\gamma} = \gamma$ is a discount factor.
The task formulation with respect to MDPRM is Markovian.
Optimal-solution guarantees of RL algorithms for MDPRMs are the same as for regular MDPs~\cite{icarte2022reward}.

\section{Problem Formulation}

To solve the SM selection problem with time-averaged constraints, we propose an RL approach that leverages RMs to handle the non-Markovian nature of the constraints
\begin{equation}
\begin{aligned}
\overline{D^{i^d}} &\le D, \quad \forall i^{d} \in \mathcal{N}^d, \\
\overline{\mu^{i^{m}}} &\ge \mu, \quad \forall i^{m} \in \mathcal{N}^{m},
\end{aligned}
\label{eq:constraints}
\end{equation}
where $D \in (0, 1)$ represents the allowed packet drop rate for the deadline-constrained traffic and $\mu < \mu^{\text{max}}$ represents the minimum throughput requirement for the constant-rate users, where $\mu^{\text{max}}$ is the maximum achievable throughput.
The key idea is to explicitly track progress toward satisfying the time-averaged constraints using an RM.

Let us first define the MDPRM.
The observable state space $S$ is continuous.
Each observation vector
\begin{equation}
    \bm{s}_t = \set{\tilde{\mu}^{\mathcal{N}^d}_t, \tilde{\mu}^{\mathcal{N}^{m}}_t, \overline{Y^{\mathcal{N}^d}_t}, \overline{Y^{\mathcal{N}^{m}}_t}, D^{\mathcal{N}^d}_t, \mu^{\mathcal{N}^{m}}_t, h^g_t \mid  g \in \mathcal{G}}
    \label{eq:state}
\end{equation}
includes the summed traffic loads $\tilde{\mu}$, packet drop rates $D$, and served throughputs $\mu$ over user group $\mathcal{N}^d$ or $\mathcal{N}^{m}$ at time $t$;
average channel conditions; and the current SM of each RU.
The observation state thus captures information about the immediate traffic load, channel conditions, and QoS performance for both user groups.

At $t$, an RL agent decides whether to put active RUs to sleep.
Let $\bm{1}_{h'}$ be the indicator of the decision to enter SM $h' \in \mathcal{H}$.
Then, the action is $\bm{a}_t = \set{\text{a}[h^g_t] \mid g \in \mathcal{G}}$, where
\begin{equation}
 \text{a}[h^g_t] = \begin{cases}
               h^g_t, & \text{if } h^g_t \neq 0, \\
               \sum_{h' \in \mathcal{H}} h'\bm{1}_{h'}(t), & \text{otherwise}.
              \end{cases}
              \label{eq:action}
\end{equation}
Therefore, the discrete action space has the size of $|A| = (H + 1)^G$, with $H$ SMs and $1$ decision to remain active.

After the agent performs an action, it receives a reward that should contain information about the energy efficiency and constraint violations.
The energy efficiency $EE \in (0, 1)$ is defined as the relative energy savings compared to the maximum power consumption when all RUs are active.
\begin{equation}
 EE_t = \frac{W^0_{t}-W_{t}}{W^0_{t}},
 \label{eq:ee}
\end{equation}
where $W^0_{t}$ and $W_{t}$ are the observed power consumptions when all RUs are active and when the RUs are in their agent-controlled states, respectively.
The drop-rate violation $\rho^d  \in (-1, 1)$ is the difference between the observed drop rate and the allowed drop rate $D$ averaged over $\mathcal{N}^d$ users, and the throughput violation $\rho^{m} \in (-1, 1)$ is the difference between the minimum required throughput $\mu$ and the served throughput averaged over $\mathcal{N}^{m}$ users.
\begin{align}
    \rho^d_t &= \overline{D^{\mathcal{N}^d}_{t}} - D, \\
    \rho^{m}_t &= \frac{1}{\mu^{\text{max}}}(\mu - \overline{\mu^{\mathcal{N}^{m}}_{t}}).
\end{align}
As the agent learns a policy that maximises the cumulative expected reward, the reward can be written as
\begin{equation}
    r^{\text{Mark}}_{t} = EE_t -
    \rho^d_t -
    \rho^{m}_t.
    \label{eq:reward_markov}
\end{equation}
The limitation of this reward function is that it is Markovian: it depends on the current state and does not account for the history of packet drops or throughput violations.

To capture the time-averaged constraints~(\ref{eq:constraints}), we use memory offered by abstract states in RMs.
The RM has access to the following propositional symbols $\MP = \set{x_{[D]}, y_{[\mu]} \mid [D], [\mu] \in \mathbb{Z}}$.
We define $[D] =\operatorname{round}(L\rho^d_t)$ and $[\mu] =\operatorname{round}(L\rho^{m}_t)$, where $\operatorname{round}(\cdot)$ rounds to the nearest integer and $L \in \mathbb{N}$  determines the granularity of the RM states.
The parameter $L$ represents the number of distinct values of the drop-rate and throughput violations that the RM can distinguish.
For modelling, we use two separate RMs: $\mathcal{M}^{j} = \set{U^j, u_0^j, F^j, \delta^j_u, \delta^j_r}$, where $j = d$ for the drop-rate constraint and $j = m$ for the throughput constraint.
We define $U^j = \set{u^j_0, u^j_1, \dots, u^j_L}$, where $u^j_0$ are the initial states and $u^j_L$ are the terminal states.
If $x_{[D]}$ is true, the transition function $\delta^d_u$ is
\begin{equation}
    \delta^d_u(u^d_l, x_{[D]}) = \begin{cases}
               u^d_{l + [D]}, & \text{if } 0 \leq l + [D] \leq L, \\
               u^{d}_{0}, & \text{if }  l + [D] < 0, \\
               u^d_L, & \text{if }  l + [D] > L.
              \end{cases}
\end{equation}
The transition function for the throughput RM $\delta^{m}_u$ is defined similarly, with $y_{[\mu]}$ instead of $x_{[D]}$.
The state-reward functions $\delta^j_r$, $j = d, m$, are defined as $\delta^j_r(u^j_l) = -\frac{l}{L}$.
These rewards are effectively non-Markovian because, for the same observable state $\bm{s}_t$, the reward can differ depending on the RM state.
The deeper the RM (the larger $L$), the more memory it has, but the more complex the learning problem for the RL agent.
The final reward received by the agent is the sum of the energy efficiency and rewards from the two RMs with depth $L$:
\begin{equation}
    r^{\text{RM:L}}_{t} = EE_t + r^{d:L}_{t} + r^{m:L}_{t}.
    \label{eq:reward_rm}
\end{equation}
The RL agent must find a policy (mapping of state $\bm{s}_t \times u^d_t \times u^{m}_t$ to action $\bm{a}_t$) that maximises the total reward over time. 

\section{Numerical Evaluation}

For the numerical evaluation, we use a system simulation tool that implements a simplified map-based ray-tracing propagation model to compute path gains at various user drops.
The system model includes RU power consumption across different SMs, switching energy costs and latencies, and wireless channel conditions for all users.

The number of users with deadline-constrained traffic $N^d$ and with constant-rate traffic $N^m$ uniformly varies from $4$ to $5$ and from $10$ to $60$, respectively. 
The traffic load $\tilde{\mu}$ is uniformly distributed between $0.1$ and $0.2$ Mbps. 
We set up four RUs and four SMs defined in~\cite{el2022energy}. 
SM1 has duration $\Delta^1=71$ $\mu$s and latency $\Delta^1_{\text{sw}} = 35.5$ $\mu$s; for SM2, $\Delta^2=1$ ms and $\Delta^2_{\text{sw}}=0.5$ ms; for SM3, $\Delta^3=10$ ms and $\Delta^3_{\text{sw}}=5$ ms; and for SM4, $\Delta^4=1$ s and $\Delta^4_{\text{sw}}=0.5$ s.
As the discrete action space is large ($5^4 = 625$), we treat it as continuous and use TD3 as the RL algorithm for learning the SM selection policy.

In the experiments, the TD3 algorithm uses the default MlpPolicy from Stable-Baselines3 (v2.2.1)~\cite{raffin2021stable}. 
Both the actor and the two critic networks consist of two fully connected hidden layers of $400$ and $300$ neurons with ReLU activations. 
The actor maps the observation vector to $4$ continuous outputs (one per RU) squashed to $[0, 1]$ via tanh and then uniformly discretised to the nearest SM level.
All networks are trained with Adam (learning rate: $0.0003$). 
The discount factor $\gamma$ is $0.2$, soft-update coefficient is $0.005$, replay buffer size is $10^6$, and mini-batch size is $256$. 
Learning starts after $500$ steps.
The policy is updated every two gradient steps.
All experiments are run on a MacBook Pro with an Apple M4 processor ($10$ cores) and $16$ GB of RAM. 
Each run lasts $5\,000$ episodes, $30$ steps per episode. 
Between episodes, the environment is reset with a new scenario (seeded) with new user numbers, user positions, traffic loads, and channel conditions and remains constant within an episode.

We test four different reward functions with the same TD3 architecture described above. 
First, we test our RM-based non-Markovian reward modelling with $L=10$ and $L= 100$. 
As one baseline, we use the Markovian reward defined in (\ref{eq:reward_markov}). 
As another baseline, we use a Lagrangian optimisation approach, a common method for constrained optimisation problems in wireless networks. 
The Lagrangian method transforms the constrained optimisation problem into an unconstrained one by introducing Lagrange multipliers for each constraint. 
The resulting problem is then solved iteratively, adjusting the multipliers based on the degree of constraint violation.
The reward remains Markovian:
\begin{equation}
    r^{\text{LO}}_{t} = EE_t - \lambda^d_t \rho^d_t - \lambda^{m}_t \rho^{m}_t,
    \label{eq:reward_lagrangian}
\end{equation}
where $\lambda^d_t$ and $\lambda^{m}_t$ are the Lagrange multipliers for the drop-rate and throughput constraints, respectively, updated with learning rates of $0.01$ per episode. 

\begin{figure}[h]
\centering
\includegraphics[width=0.5\textwidth]{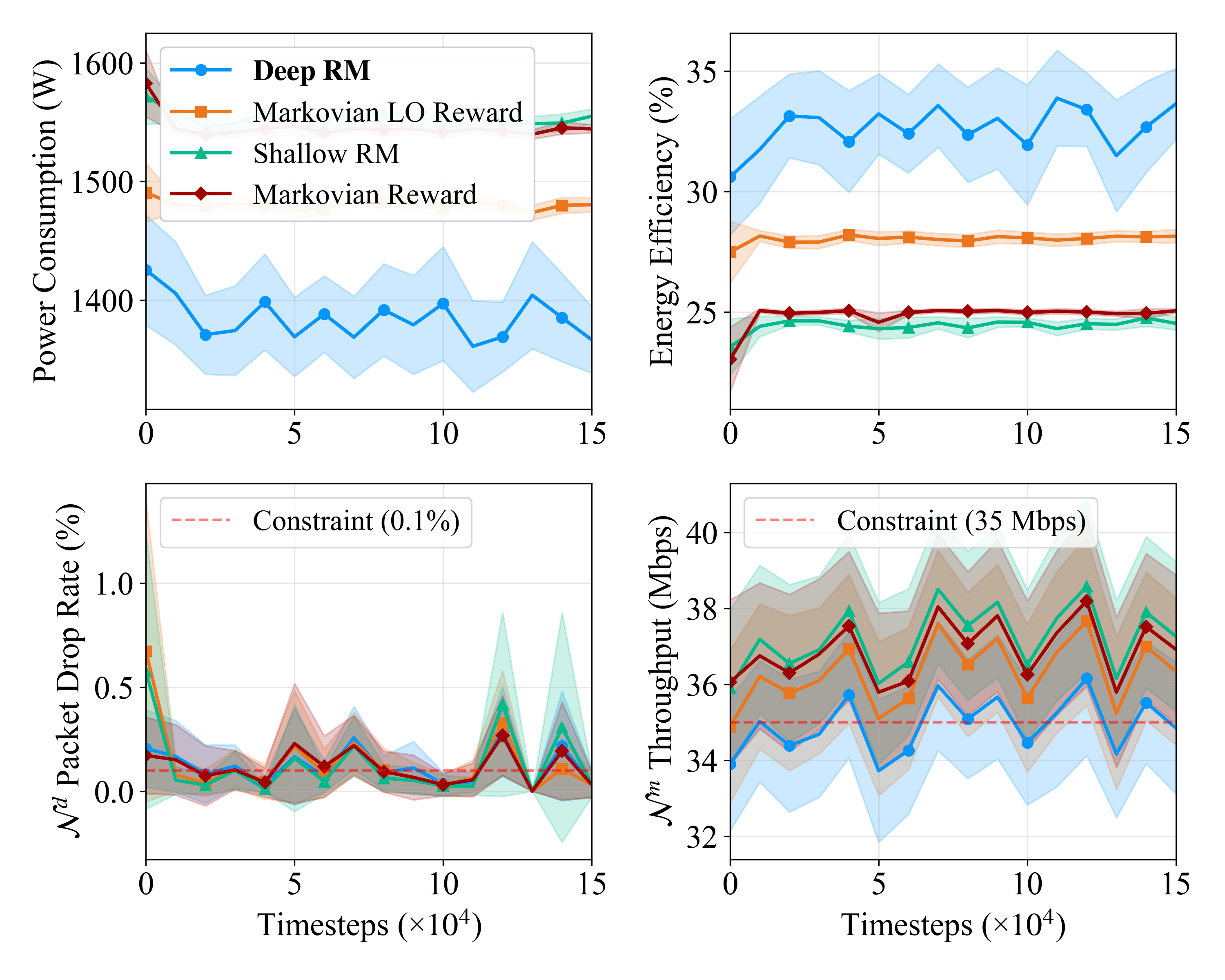}
\caption{Power consumption, energy efficiency, and constraint satisfaction for the TD3 RL agents with the rewards from the deep RM (\ref{eq:reward_rm}) with $L = 100$ and shallow RM (\ref{eq:reward_rm}) with $L = 10$, Markovian reward (\ref{eq:reward_markov}), and Lagrangian-optimised (LO) Markovian reward (\ref{eq:reward_lagrangian}).
Shaded regions represent $95\%$ confidence intervals.}
\label{fig:comparison}
\end{figure}

\begin{figure}[h]
\centering
\includegraphics[width=0.5\textwidth]{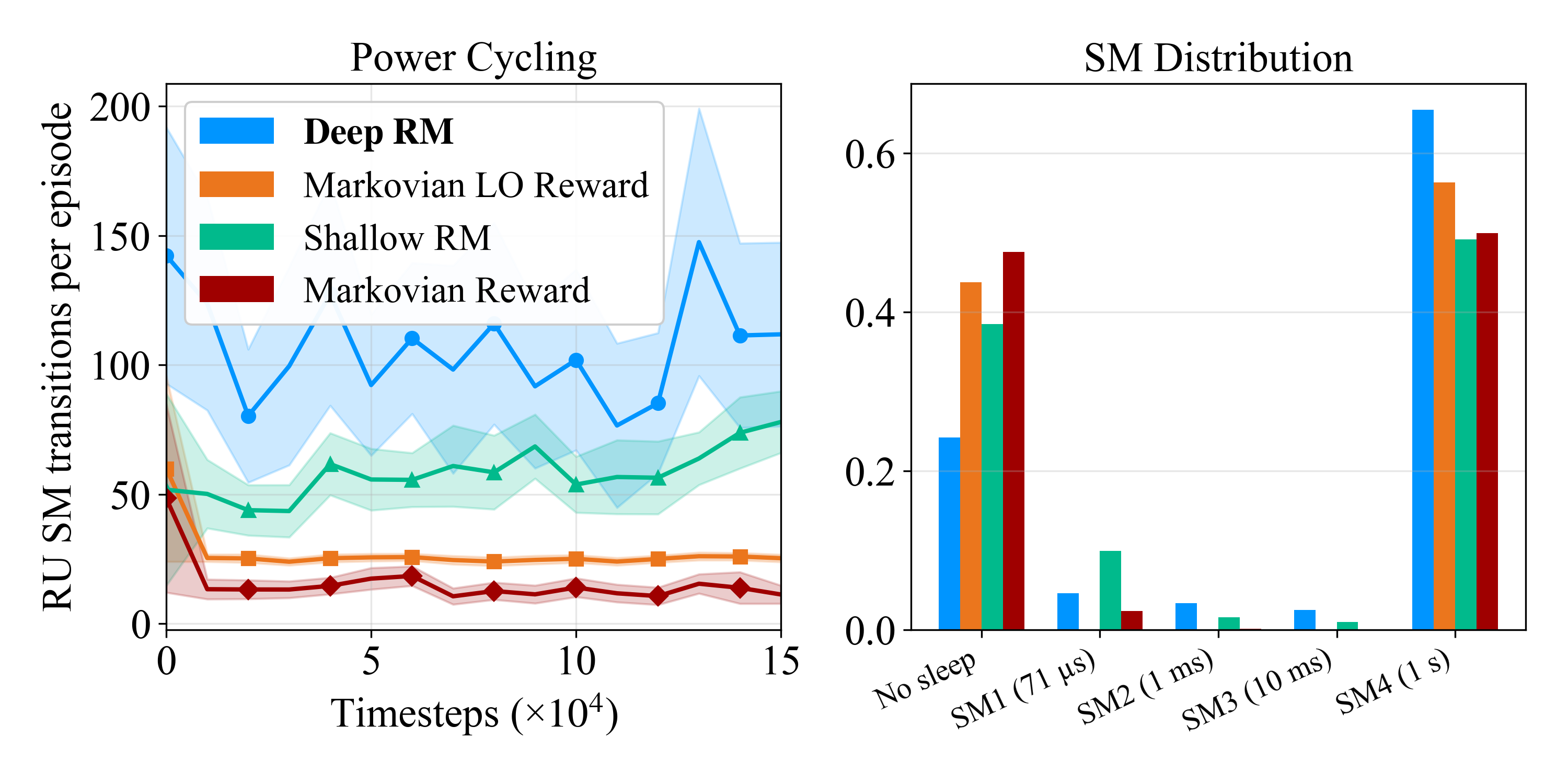}
\caption{Power cycling and converged sleep mode (SM) distribution.}
\label{fig:power-cycling}
\end{figure}

\begin{figure}[h]
\centering
\includegraphics[width=0.5\textwidth]{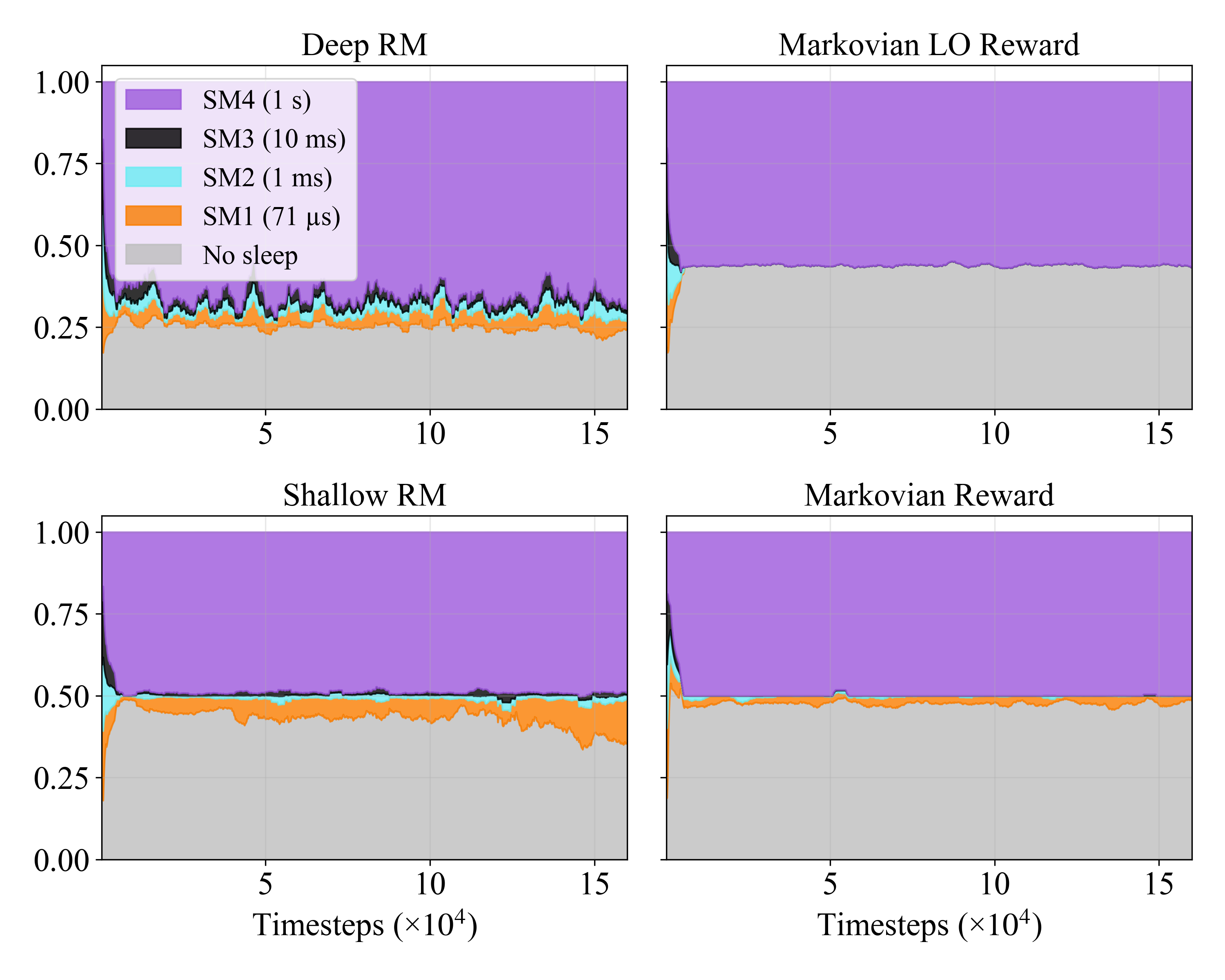}
\caption{Policy analysis via SM distribution of each agent.}
\label{fig:sm-dist}
\end{figure}

\section{Discussion and Conclusion}

The experimental comparison of power consumption, energy efficiency (EE), and constraint satisfaction is shown in Fig.~\ref{fig:comparison}. 
The results indicate that the deep-RM-based agent achieves the highest EE while operating close to the constraint boundary. 
By contrast, the shallow-RM-based agent is more conservative, even compared with the LO-reward-based agent. 
This suggests that additional RM memory is beneficial: with a deeper RM, the agent accumulates a richer history of past violations and can therefore learn a more nuanced policy. 
In particular, it can strategically use the available ``violation budget'', allowing temporary violations in difficult scenarios to improve long-term EE. Hence, the RM depth $L$ is a key design parameter.

This behavior is further supported by the power-cycling results in Fig.~\ref{fig:power-cycling} and the SM distribution in Fig.~\ref{fig:sm-dist}. 
Among all the agents, the deep-RM-based agent changes the RU SMs most often, indicating higher policy adaptability. 
The Markovian-reward-based agent changes SMs least often, followed by the LO-reward-based agent and then the shallow-RM-based agent. 
Intuitively, a high EE is achieved by the agents that keep RUs asleep for a large fraction of time. 
The deep-RM-based agent is mostly in the longest SM (SM4), while still using SM1--SM3 when needed. 
Its large variation across episodes indicates strong scenario-dependent adaptation. 
In contrast, the Markovian-reward-based agents tend to adopt a simpler bimodal behavior: either SM4 or active mode, because their policy is, by design, optimised for immediate constraint satisfaction.

Overall, the numerical results show that non-Markovian reward modelling with sufficiently deep RMs improves the trade-off between EE and long-term QoS compliance. 
These findings suggest that RMs are a promising abstraction for embedding temporal constraint information into RL-based network-control policies.

\section{Acknowledgements}

We thank Elliot Gestrin, Windy Phung, and Farid Musayev for their constructive feedback and suggestions.

\bibliographystyle{IEEEtran}
\bibliography{references}

\end{document}